\documentclass[11pt]{article}

\usepackage{acl}
\usepackage{enumitem}
\usepackage[most]{tcolorbox}
\usepackage{times}
\usepackage{latexsym}
\usepackage[T1]{fontenc}
\usepackage[utf8]{inputenc}
\usepackage{microtype}
\usepackage{inconsolata}
\usepackage{graphicx}
\usepackage{amsmath}
\usepackage{booktabs}
\usepackage{multirow}
\usepackage{algorithm}
\usepackage{algorithmic}
\usepackage{hyperref}
\usepackage{tcolorbox}
\tcbuselibrary{breakable}

\title{RAM-SD: Retrieval-Augmented Multi-agent \\ framework for Sarcasm Detection}

\author{
Ziyang Zhou\thanks{* Equal contribution.},
Ziqi Liu\footnotemark[1],
Yan Wang, Yiming Lin \and Yangbin Chen \\
Xi'an Jiaotong--Liverpool University \\
\texttt{\{ziyang.zhou22, ziqi.liu22, yan.wang2202, yiming.lin21, yangbin.chen\}@xjtlu.edu.cn}
}
  
\begin{document}
\maketitle

\begin{abstract}
Sarcasm detection remains a significant challenge due to its reliance on nuanced contextual understanding, world knowledge, and multi-faceted linguistic cues that vary substantially across different sarcastic expressions. Existing approaches, from fine-tuned transformers to large language models, apply a uniform reasoning strategy to all inputs, struggling to address the diverse analytical demands of sarcasm. These demands range from modeling contextual expectation violations to requiring external knowledge grounding or recognizing specific rhetorical patterns. To address this limitation, we introduce \textbf{RAM-SD}, a \textbf{R}etrieval-\textbf{A}ugmented \textbf{M}ulti-Agent framework for \textbf{S}arcasm \textbf{D}etection. The framework operates through four stages: (1) contextual retrieval grounds the query in both sarcastic and non-sarcastic exemplars; (2) a meta-planner classifies the sarcasm type and selects an optimal reasoning plan from a predefined set; (3) an ensemble of specialized agents performs complementary, multi-view analysis ; and (4) an integrator synthesizes these analyses into a final, interpretable judgment with a natural language explanation. Evaluated on four standard benchmarks, RAM-SD achieves a state-of-the-art Macro-F1 of 77.74\%, outperforming the strong GPT-4o+CoC baseline by \textbf{7.01} points. Our framework not only sets a new performance benchmark but also provides transparent and interpretable reasoning traces, illuminating the cognitive processes behind sarcasm comprehension.

\end{abstract}

\section{Introduction}

Sarcasm detection remains a formidable challenge in natural language processing due to the intricate interplay of context, world knowledge, and multi-faceted linguistic cues required for its comprehension \citep{joshi2017automatic}. The ability to accurately identify sarcastic intent is crucial for developing more robust human-computer interaction systems and for enabling deeper analysis of social media content, from public opinion mining to user modeling. However, the sheer diversity of sarcastic expression, which ranges from simple verbal irony to complex, culturally-grounded statements, presents a significant hurdle for automated systems \citep{riloff2013sarcasm}. Early research sought to capture sarcastic signals through feature engineering \citep{davidov2010semi} and conventional deep learning architectures \citep{ghosh2017role}. More recently, the advent of Pre-trained Language Models such as BERT \citep{devlin2019bert}, and the subsequent emergence of LLMs have established new performance benchmarks, showcasing remarkable capabilities in understanding nuanced text through prompting strategies like chain-of-thought \citep{wei2022chain}.

Despite recent advances, state-of-the-art approaches are constrained by a uniform reasoning strategy ill-suited for the diverse nature of sarcasm. Different sarcastic expressions demand distinct analytical methods, such as contextual reasoning for expectation violations \citep{grice1975logic}, cultural grounding for knowledge-dependent irony \citep{li2024cross}, and pattern recognition for rhetorical devices \citep{gibbs2000irony}. This monolithic approach is further challenged by cognitive research indicating that human sarcasm processing involves parallel neural networks \citep{rapp2012lateral}, which suggests that single, unified models are inherently insufficient for this task. Consequently, even powerful LLMs lack the adaptivity to dynamically tailor their analysis, leading to inconsistent performance across varied types of sarcasm.

To overcome these limitations, we draw inspiration from the parallelism in human cognition to reframe sarcasm detection as a dynamic, adaptive reasoning process rather than a static classification task. We introduce RAM-SD, a Retrieval-Augmented Multi-Agent framework designed to instantiate this vision. Central to our approach is a Meta-Planner that first analyzes the input text alongside retrieved contextual exemplars. Based on this analysis, it dynamically selects and dispatches a tailored ensemble of specialized agents to perform multi-faceted reasoning. The complementary analyses from these agents are then synthesized into a final, coherent prediction accompanied by an interpretable explanation.

This paper presents a comprehensive evaluation of RAM-SD, demonstrating its effectiveness and architectural integrity. Our primary contributions are as follows:
\begin{itemize}
\item We propose RAM-SD, a novel cognitively-inspired framework where a meta-planner orchestrates a retrieval-augmented multi-agent system to perform adaptive sarcasm analysis.
\item The framework achieves state-of-the-art performance on four benchmarks, with an 77.74\% average Macro-F1 score that outperforms GPT-4o+CoC by 7.01 points.
\item It offers enhanced interpretability by generating structured reasoning traces that make the model's multi-stage analytical process transparent.
\end{itemize}

To the best of our knowledge, this is the first systematic instantiation of a retrieval-augmented, meta-planning multi-agent paradigm for sarcasm detection. The design improves interpretability via evidence-aligned rationales and reduces single-model bias through specialized agents and cross-checks, offering a principled pathway to reliable context-grounded reasoning and a promising direction for future work.

\section{Related Work}
\subsection{Traditional Feature Engineering Approaches}
Early computational approaches to sarcasm detection were characterized by their reliance on handcrafted features designed to capture explicit signals of irony. A seminal line of inquiry focused on the principle of incongruity, exemplified by the contrast hypothesis which identifies sarcasm through the co-occurrence of positive sentiment and a negative situation \citep{riloff2013sarcasm}. Following this, researchers broadened their investigation to include a diverse set of linguistic markers. These included lexical patterns such as specific n-grams and interjections \citep{davidov2010semi}, overt typographical cues like exclamation marks and hashtags \citep{liebrecht2013perfect}, and various syntactic structures. The scope was further extended to incorporate pragmatic information, with studies demonstrating the utility of conversational context and author-specific attributes for more accurate classification \citep{bamman2015contextualized, joshi2015automatic}. These methods, however, relied heavily on manual feature engineering and struggled to capture subtle sarcasm that required deep contextual or world knowledge, highlighting the need for architectures capable of learning complex representations directly from raw text.

\subsection{The Deep Learning Revolution}
The advent of deep learning instigated a paradigm shift from feature engineering to automated representation learning. This transition allowed models to learn salient features directly from raw text, significantly improving their ability to capture complex linguistic phenomena. Early applications in this domain included Recurrent Neural Networks (LSTMs), which proved effective for modeling the sequential dependencies inherent in conversational data \citep{ghosh2017role}, and Convolutional Neural Networks (CNNs), which were utilized to detect local, position-invariant patterns indicative of sarcasm \citep{misra2016using}. The field was fundamentally transformed by the arrival of pre-trained language models based on the Transformer architecture, such as BERT \citep{devlin2019bert}. By leveraging deep contextualized embeddings, these models set new standards for performance and became the predominant approach for the task \citep{potamias2020transformer}. While these pre-trained models excelled at contextual understanding, their reliance on task-specific fine-tuning and their fixed architectures limited their adaptability, paving the way for more general, massively-scaled language models with emergent reasoning abilities.

\subsection{Large Language Model Approaches}
The current frontier in sarcasm detection is defined by the capabilities of LLMs such as the GPT series \citep{brown2020language}. Comprehensive evaluations have benchmarked these models, confirming their state-of-the-art performance while also revealing challenges related to prompt sensitivity and the comprehension of cultural nuances \citep{miranskyy2023sarcasm, lou2024sarcasmbench, liu2025sevade}. To elicit this performance, researchers employ a range of strategies, from few-shot in-context learning to advanced prompting techniques like Chain-of-Thought, with recent work investigating the nature and efficacy of such step-by-step reasoning processes for sarcasm \citep{wei2022chain,yao2025sarcasm}. While these methods enhance the reasoning of a single model, they do not alter its fundamental monolithic architecture. This uniform approach is inherently limited when faced with a cognitively complex task like sarcasm, which requires an orchestration of diverse analytical skills such as contextual reasoning and world knowledge \citep{grice1975logic, rapp2012lateral}. Motivated by this architectural mismatch, our work proposes a departure from the single-model paradigm towards a framework built on adaptive, specialized reasoning.

\begin{figure*}[t]
    \makebox[\textwidth][c]{
        \includegraphics[width=1.15\textwidth]{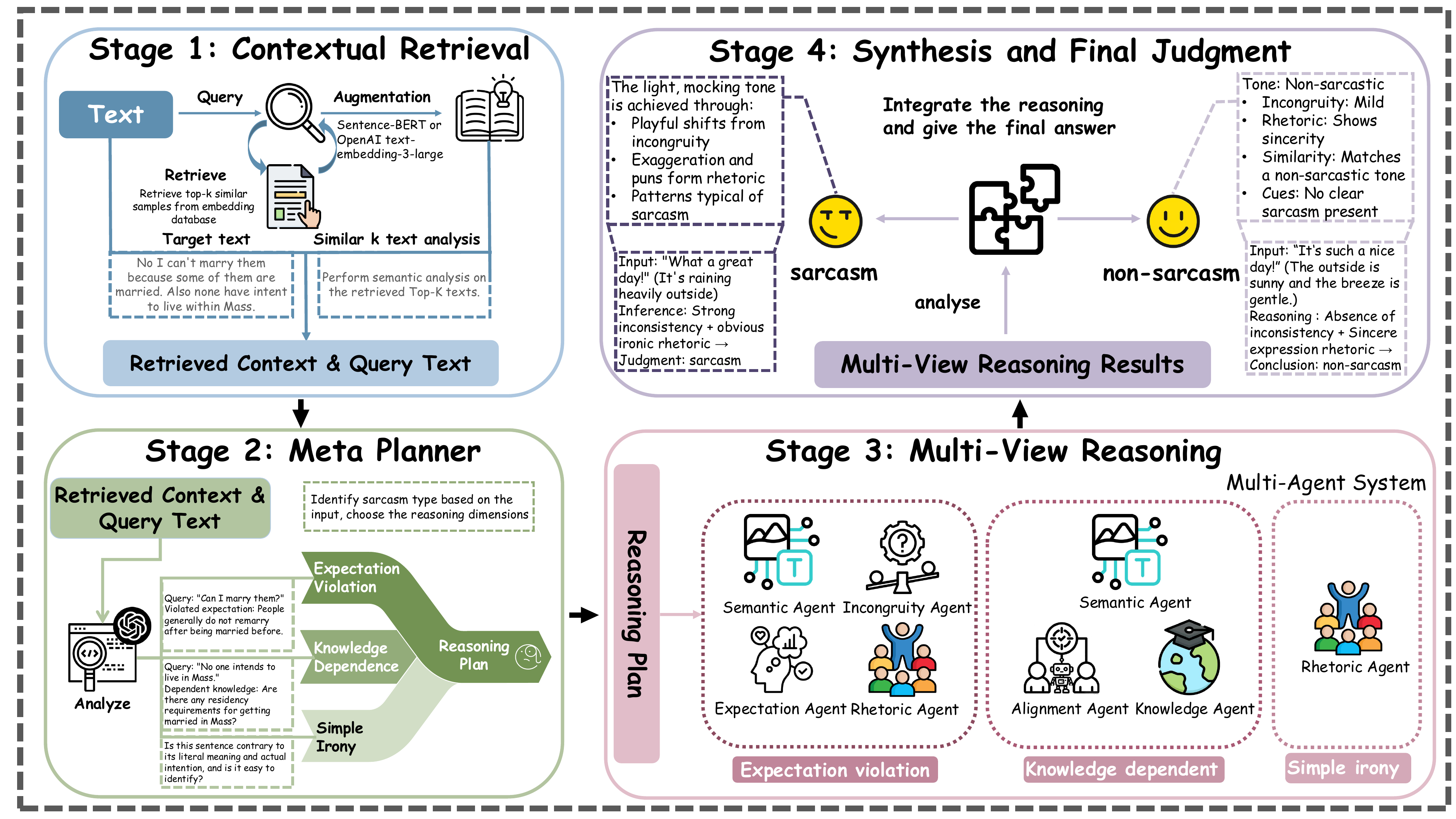}
    }
    \caption{The four-stage architecture of the RAM-SD framework. Stage 1 grounds the query with retrieved sarcastic and non-sarcastic exemplars. This informs the Stage 2 meta-planner which selects a tailored reasoning plan. Stage 3 executes this plan with an ensemble of specialized agents. The agents' findings are synthesized in Stage 4 to produce a final judgment and explanation.}
    \label{fig:architecture}
\end{figure*}

\section{Methodology}

We introduce \textbf{RAM-SD}, a Retrieval-Augmented Multi-Agent framework for Sarcasm Detection. As illustrated in Figure~\ref{fig:architecture}, RAM-SD operates through a four-stage cognitive pipeline: (1) Contextual Retrieval, (2) Retrieval-Augmented Meta-Planning, (3) Agent-based Multi-faceted Reasoning, and (4) Synthesis and Final Judgment.

\subsection{Stage 1: Contextual Retrieval}
This module grounds the analysis in relevant background knowledge. For a given query text $T_q$, we retrieve semantically relevant exemplars from a pre-compiled knowledge base $\mathcal{D}=\{(T_i, y_i)\}_{i=1}^M$.

\paragraph{Dual-Subset Retrieval.} All texts are encoded into vectors using OpenAI’s \texttt{text-embedding-3-large}. To mitigate label imbalance and enhance contrast, we partition the vectorized knowledge base into sarcastic $\mathcal{D}_{\text{vec}}^{\text{sarc}}$ and non-sarcastic $\mathcal{D}_{\text{vec}}^{\text{non}}$ subsets. We then retrieve the top-$k$ exemplars from each subset based on similarity to $T_q$, forming a balanced retrieved set $\mathcal{C}_{\text{ret}} = \mathcal{C}_{k}^{\text{sarc}} \cup \mathcal{C}_{k}^{\text{non}}$ of size $2k$.

\paragraph{Rationale-Enhanced Contextualization.} To transform retrieved instances into actionable insights, we use an LLM-based analyzer, $\text{LLM}_{\text{rat}}$, to generate a concise rationale $r_i$ for each exemplar $(T_i, y_i) \in \mathcal{C}_{\text{ret}}$. The rationale explains \emph{why} the text is sarcastic or not by identifying cues like hyperbole, sentiment-context mismatch, or knowledge contradictions. This creates a rationale-augmented context set, $\mathcal{C}_{\text{aug}} = \{(T_i, y_i, r_i)\}_{i=1}^{2k}$, which provides both examples and the underlying reasoning for downstream modules.

\subsection{Stage 2: Retrieval-Augmented Meta-Planning}
This module acts as a dynamic planner, selecting a tailored reasoning strategy for the query $T_q$. The planner $\text{LLM}_{\text{plan}}$ analyzes $T_q$ by referencing the rationale-augmented exemplars in $\mathcal{C}_{\text{aug}}$, inferring potential context and background information. It produces two outputs: (1) the selected reasoning plan $P^{*}$, and (2) a contextual analysis $O_{\text{plan}}$ that synthesizes inferred situational context based on $\mathcal{C}_{\text{aug}}$:
\[
(P^{*}, O_{\text{plan}}) = \text{LLM}_{\text{plan}}(T_q, \mathcal{C}_{\text{aug}})
\]
where $P^{*} = \arg\max_{P_i \in \mathcal{P}} P(P_i \mid T_q, \mathcal{C}_{\text{aug}})$ and $O_{\text{plan}}$ contains inferred context that guides subsequent agent reasoning. Both $P^{*}$ and $O_{\text{plan}}$ are passed to Stage~3.

\paragraph{Pre-defined Reasoning Plans.} We define three plans targeting distinct sarcasm archetypes:
\begin{itemize}[noitemsep]
    \item \textbf{Expectation Violation Plan ($\mathcal{P}_{\text{EV}}$):} Targets sarcasm arising from pragmatic incongruity. It deploys agents focused on semantics, expectations, incongruity, and rhetoric: $\{A_{\text{sem}}, A_{\text{exp}}, A_{\text{incon}}, A_{\text{rhet}}\}$.
    \item \textbf{Knowledge-Dependent Plan ($\mathcal{P}_{\text{KD}}$):} For sarcasm requiring external world knowledge. It uses agents for semantics, knowledge grounding, alignment, and rhetoric: $\{A_{\text{sem}}, A_{\text{know}}, A_{\text{align}}, A_{\text{rhet}}\}$.
    \item \textbf{Simple Irony Plan ($\mathcal{P}_{\text{SI}}$):} A lightweight plan for overt irony, primarily relying on the rhetoric agent $\{A_{\text{rhet}}\}$ for efficient detection,  with optional $\{A_{\text{sem}}, A_{\text{incon}}\}$ support for ambiguous cases.
\end{itemize}

The output of this module is the selected plan $P^{*}$, which dictates the precise set of agents and the reasoning workflow for the next stage.

\subsection{Stage 3: Agent-based Multi-View Reasoning }
Following the meta-planner's directive, the selected agent ensemble $P^{*}$ analyzes $T_q$ from complementary perspectives. Each agent receives $T_q$, the contextual analysis $O_{\text{plan}}$, and the rationale-augmented exemplars $\mathcal{C}_{\text{aug}}$ as input. Agents operate through structured prompting, producing interpretable reasoning outputs:
\begin{itemize}[noitemsep]
\item \textbf{Semantic Agent ($A_{\text{sem}}$):} Provides a literal interpretation of the text, identifying sentiment polarity and assessing surface-level coherence.
\item \textbf{Rhetoric Agent ($A_{\text{rhet}}$):} Detects rhetorical signals such as hyperbole, understatement, or rhetorical questions, and qualitatively estimates their strength and intent through prompt-guided reasoning.
\item \textbf{Expectation Agent ($A_{\text{exp}}$):} Examines the alignment between the situation implied by $T_q$ and general world knowledge. Although not based on explicit statistical divergence, its reasoning approximates a measure conceptually akin to a deviation between expected and observed contexts.
\item \textbf{Knowledge Agent ($A_{\text{know}}$):} Identifies key entities or events and retrieves relevant factual or cultural information through prompt-based grounding. The resulting summaries enrich contextual understanding.
\item \textbf{Alignment Agent ($A_{\text{align}}$):} Evaluates whether the semantics of $T_q$ are more consistent with sarcastic or non-sarcastic patterns in $\mathcal{C}_{\text{aug}}$, guided by $O_{\text{plan}}$.
\item \textbf{Incongruity Agent ($A_{\text{incon}}$):} Synthesizes inconsistencies among semantic, rhetorical, and contextual dimensions into an overall incongruity judgment.
\end{itemize}
Each agent $A_j \in P^{*}$ produces a reasoning output $R_j = A_j(T_q, O_{\text{plan}}, \mathcal{C}_{\text{aug}})$. The collective outputs form a structured reasoning trace $\mathcal{R}_{P^{*}} = \{R_j \mid A_j \in P^{*}\}$, serving as the foundation for final synthesis.

\subsection{Stage 4: Synthesis and Final Judgment}
In the final stage, all agent outputs from $\mathcal{R}_{P^{*}}$ are first aggregated by an Integrator component, which synthesizes the diverse analytical perspectives into a coherent evidence summary. 
Subsequently, this integrated reasoning trace is evaluated by the Judger model, $\text{LLM}_{\text{judge}}$, to produce the final verdict. 
The Judger receives a comprehensive prompt $P_{\text{final}} = T_{\text{judge}}(T_q, \mathcal{R}_{P^{*}})$, which consolidates the original query $T_q$ and the multi-faceted reasoning trace $\mathcal{R}_{P^{*}}$. 
This design ensures that the final prediction is grounded in the aggregated reasoning evidence while maintaining interpretability and consistency across agent decisions.

\paragraph{Prediction and Explanation.}
The Judger performs two crucial tasks. 
First, it predicts the final probability of the text being sarcastic:
\[
p(y=\textit{sarcasm} \mid T_q, \mathcal{R}_{P^{*}}) = \text{LLM}_{\text{judge}}(P_{\text{final}})
\]
The final label $y_{\text{pred}}$ is determined by thresholding this probability at 0.5. 
Second, and critically for interpretability, it generates a natural-language explanation based on $P_{\text{final}}$, which is not a mere concatenation of agent outputs but a coherent synthesis that highlights the most salient evidence and logical steps leading to the conclusion. 
This dual output provides both a quantitative prediction and a qualitative justification, completing the RAM-SD pipeline by delivering a robust, transparent, and contextually aware sarcasm detection judgment.

\section{Experiments}
We conduct comprehensive experiments to evaluate the effectiveness of RAM-SD across diverse sarcasm detection benchmarks. Our evaluation focuses on: (1) Overall performance comparison with state-of-the-art baselines, (2) Ablation studies to validate the contribution of key architectural components, (3) Sensitivity analysis of the retrieval parameter, (4) Analysis of inference efficiency, and (5) Comprehensive evaluation of the framework's interpretability and limitations through quality assessment and error pattern analysis.

\subsection{Experimental Setup}

\subsubsection{Datasets}

We evaluate RAM-SD on four widely-used sarcasm detection benchmarks covering diverse linguistic styles and contexts. Table~\ref{tab:datasets} summarizes the dataset statistics.

\renewcommand{\arraystretch}{0.9} 
\setlength{\tabcolsep}{4pt}      
\begin{table}[!htp]
\centering
\caption{Overview of the benchmark datasets used for evaluating sarcasm detection.}
\label{tab:datasets}
\begin{tabular}{lcccc}
\toprule
Dataset & Train & Test & Context & Source \\
\midrule
IAC-V1 & 1,595 & 320 & No & Reddit \\
IAC-V2 & 5,216 & 1042 & No & Reddit \\
MUSTARD & 552 & 138 & Yes & TV Shows \\
SemEval & 3,634 & 784 & No & Twitter \\
\bottomrule
\end{tabular}
\end{table}

\textbf{IAC-V1/V2} \citep{oraby2016creating} contain political debate posts from Reddit's \textit{r/politics} forum, where sarcasm often manifests through expectation violations and implicit critique. \textbf{MUSTARD} \citep{ghosh2020mustard} provides conversational utterances from TV show dialogues with contextual background, requiring understanding of character relationships and situational awareness. \textbf{SemEval-2018 Task 3} \citep{vanhee2018semeval} comprises Twitter posts where sarcasm frequently relies on simple irony and rhetorical devices within character-limited expressions.
\subsubsection{Baselines}
We compare RAM-SD against representative methods from three categories:

\noindent \textbf{Deep Learning Models:} \textbf{MIARN} \citep{tay2018reasoning} employs multi-interactive attention networks; \textbf{SAWS} \citep{pan2020saws} leverages sentiment-aware word selection; \textbf{DC-Net} \citep{liu2021dual} utilizes dual-channel networks for multi-view reasoning.

\noindent \textbf{Fine-tuned PLMs:} \textbf{BERT-base} \citep{devlin2019bert} and \textbf{RoBERTa-base} \citep{liu2019roberta} fine-tuned on each target dataset using standard classification protocols.

\noindent \textbf{LLM-based Methods:} \textbf{LLM-based Methods:} We compare RAM-SD against several state-of-the-art LLM-based methods. Our baselines include prompting strategies from SarcasmCue \citep{yao2025sarcasm}, such as GPT-4o with Chain of Contradiction, Graph of Cues, and Bagging of Cues prompting. We also incorporate \textbf{IDADP} \citep{yi2025irony} and \textbf{CAF-I} \citep{liu2025caf} frameworks.

\subsubsection{Implementation Details}
We use OpenAI's \texttt{text-embedding-3-large} (3072 dimensions) for text embeddings in the Contextual Retrieval stage. The Meta-Planner, all specialized agents, and the Judger are implemented using GPT-4o with temperature=0.1 and max\_tokens=512. Vector similarity search employs FAISS \citep{johnson2019billion} with cosine similarity after L2 normalization. Unless otherwise specified, we set $k=3$ for dual-subset retrieval, yielding $|\mathcal{C}_{\text{ret}}| =6$ balanced exemplars.

\subsection{Main Results}

Table~\ref{tab:results} presents the performance comparison across all datasets. RAM-SD achieves state-of-the-art performance with an average Macro-F1 of 77.74\% and accuracy of 77.65\%, substantially outperforming all baseline methods.

\begin{table*}[!htp]
\centering
\caption{Overall performance comparison across four benchmark datasets. All LLM strategies are zero-shot. Acc. denotes Accuracy and Ma-F1 signifies Macro-F1. Best results are in bold, second-best are \underline{underlined}.}
\label{tab:results}
\small
\begin{tabular*}{\textwidth}{@{\extracolsep{\fill}}l|cc|cc|cc|cc|cc}
\toprule
\multirow{2}{*}{\textbf{Method}} & \multicolumn{2}{c|}{\textbf{IAC-V1}} & \multicolumn{2}{c|}{\textbf{IAC-V2}} & \multicolumn{2}{c|}{\textbf{MUSTARD}} & \multicolumn{2}{c|}{\textbf{SemEval 2018}} & \multicolumn{2}{c}{\textbf{Avg.}}\\
\cmidrule(lr){2-3} \cmidrule(lr){4-5} \cmidrule(lr){6-7} \cmidrule(lr){8-9} \cmidrule(lr){10-11}
& Acc. & Ma-F1 & Acc. & Ma-F1 & Acc. & Ma-F1 & Acc. & Ma-F1 & Acc. & Ma-F1\\
\midrule
MIARN & 63.21 & 63.18 & 72.75 & 72.75 & 64.60 & 63.90 & 68.50 & 67.80 & 67.26 & 66.91 \\
SAWS & 66.13 & 65.60 & 76.20 & 76.20 & 69.71 & 70.95 & 69.90 & 68.90 & 70.48 & 70.41\\
DC-Net & 66.50 & 66.40 & 78.00 & 77.90 & \underline{71.28} & \underline{71.43} & 76.30 & 76.70 & 73.02 & 73.11 \\
\midrule
BERT & 65.30 & 65.20 & 76.40 & 76.20 & 64.30 & 64.30 & 69.90 & 68.40 & 68.97 & 68.52 \\
RoBERTa & 70.10 & 69.90 & 76.60 & 76.70 & 66.10 & 66.00 & 70.20 & 69.10 & 70.75 & 70.42\\
\midrule
GPT-4o & 70.63 & 70.05 & 73.03 & 71.99 & 67.24 & 65.79 & 64.03 & 63.17 & 68.73 & 67.75\\
GPT-4o+CoT & 61.56 & 58.49 & 58.83 & 56.42 & 58.92 & 51.99 & 58.11 & 55.76 & 59.36 & 55.67 \\
GPT-4o+CoC & 72.19 & 71.52 & 73.36 & 72.31 & 69.42 & 68.48 & 70.79 & 70.60 & 71.44 & 70.73\\
GPT-4o+GoC & 69.84 & 69.20 & 72.47 & 71.60 & 68.10 & 67.42 & 65.91 & 65.03 & 69.58 & 68.31\\
GPT-4o+BoC & 70.32 & 70.11 & 73.12 & 72.08 & 67.53 & 66.12 & 64.88 & 63.92 & 68.96 & 68.06\\
CAF-I & \underline{73.75} & \underline{73.71} & 77.80 & 76.82 & \textbf{75.21} & \textbf{74.73} & 80.73 & 79.99 & \underline{76.89} & \underline{76.31}\\
IDADP & 65.84 & 67.13 & 70.32 & 69.73 & 67.44 & 67.37 & 65.31 & 65.28 & 67.22 & 67.37\\
\midrule
\textbf{RAM-SD w/o RAG} & 71.25 & 71.23 & \underline{78.69} & \underline{78.68} & 69.38 & 69.32 & \underline{80.94} & \underline{80.90} & 75.07 & 75.05 \\
\textbf{RAM-SD} & \textbf{74.81} & \textbf{74.45} & \textbf{80.13} & \textbf{80.11} & 69.38 & 69.32 & \textbf{86.30} & \textbf{87.10} & \textbf{77.65} & \textbf{77.74} \\
Improv. & 1.06$\uparrow$ & 0.74$\uparrow$ & 2.13$\uparrow$ & 2.21$\uparrow$ & - & - & 5.36$\uparrow$ & 6.20$\uparrow$ & 0.27$\uparrow$ & 0.94$\uparrow$ \\
p-val. & $5.82\text{e}^{-3}$ & $1.91\text{e}^{-4}$ & $7.04\text{e}^{-3}$ & $3.36\text{e}^{-3}$ & $9.15\text{e}^{-4}$ & $2.28\text{e}^{-3}$ & $6.77\text{e}^{-3}$ & $4.59\text{e}^{-4}$ & $8.13\text{e}^{-3}$ & $3.51\text{e}^{-3}$ \\
\bottomrule
\end{tabular*}
\end{table*}

\textbf{Performance Analysis:} RAM-SD achieves consistent improvements across all datasets, with particularly strong performance on SemEval (86.30\% Accuracy) where contextual information enables better knowledge grounding. The system outperforms the strongest baseline GPT-4o+CoC by 7.01 points on average (70.73\% vs. 77.74\%).

\subsection{Ablation Study}

\begin{table}[ht]
\centering
\caption{Ablation Study Results (Average Macro-F1\% across Four Datasets)}
\label{tab:ablation}
\small
\resizebox{\columnwidth}{!}{%
\begin{tabular}{@{\hspace{0.3cm}}lcc@{\hspace{0.3cm}}}
\toprule
Configuration & Ma-F1 & $\Delta$Ma-F1 \\
\midrule
\textbf{Full RAM-SD System} & \textbf{77.74} $\pm$ 0.8 & - \\
\midrule
w/o Contextual Retrieval (Stage 1) & 75.05 $\pm$ 0.9 & -2.69 \\
w/o Meta-Planner (Stage 2) & 74.47 $\pm$ 1.1 & -3.27 \\
w/o Rhetoric Agent & 75.17 $\pm$ 0.7 & -2.57 \\
w/o Knowledge Agent & 75.31 $\pm$ 0.8 & -2.43 \\
w/o Incongruity Agent & 74.65 $\pm$ 0.9 & -3.09 \\
\bottomrule
\end{tabular}
} 
\end{table}

To validate the architectural design of RAM-SD, we conduct ablation studies isolating the contribution of key components. Table~\ref{tab:ablation} presents results averaged across all four datasets.

The most substantial performance degradation arises from removing the \textbf{Meta-Planner (Stage 2)}, which causes a \textbf{3.27\%} Ma-F1 drop. This highlights that adaptive reasoning-plan selection is crucial for handling the diverse manifestations of sarcasm. The \textbf{Incongruity Agent} follows with a \textbf{3.09\%} decrease, underscoring the necessity of detecting semantic-pragmatic inconsistencies. Eliminating the \textbf{Contextual Retrieval} leads to a \textbf{2.69\%} reduction, confirming that grounding analysis in contextual exemplars remains vital for recognizing nuanced sarcastic patterns. The \textbf{Rhetoric Agent} yields a \textbf{2.57\%} decline, showing that rhetorical-cue modeling provides complementary interpretive depth. Finally, removing the \textbf{Knowledge Agent} results in a \textbf{2.43\%} drop, suggesting that world knowledge enhances contextual understanding without dominating the reasoning process.

\subsection{Retrieval Parameter Sensitivity}

\begin{figure}[ht]
    \centering
    \includegraphics[width=0.50\textwidth]{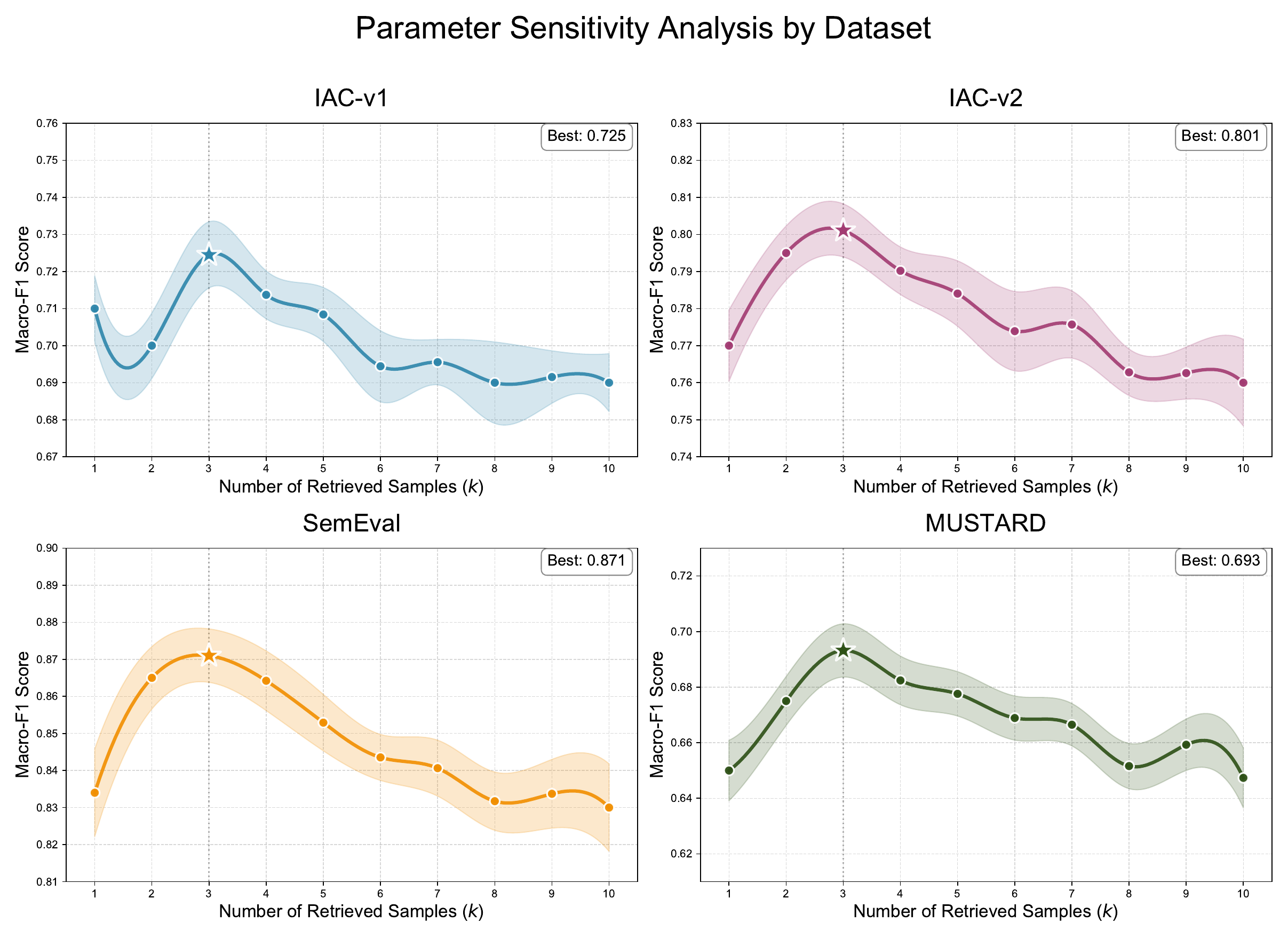}
    \caption{Impact of retrieval parameter $k$ on IAC-V1,V2, SemEval 2018 and MUSTARD datasets. Performance peaks at $k=3$ and plateaus for larger values.}
    \label{fig:sensitivity}
\end{figure}

We analyze the sensitivity of RAM-SD to the retrieval parameter $k$, which controls the number of exemplars retrieved from each subset (sarcastic and non-sarcastic). Figure~\ref{fig:sensitivity} shows Macro-F1 performance on IAC-V1 as $k$ varies from 1 to 10.

Performance improves as $k$ increases from 1 to 3, confirming that richer contextual grounding benefits downstream reasoning. However, performance plateaus at $k=3$ and slightly declines for $k>5$, suggesting that excessive exemplars introduce noise from less relevant samples. This validates our default choice of $k=3$ as balancing contextual richness with focused relevance.





\subsection{Inference Efficiency Analysis}

Beyond accuracy improvements, the meta-planner provides computational efficiency through adaptive agent selection. Table~\ref{tab:efficiency} presents inference time breakdown across the four-stage pipeline.

\begin{table}[!ht]
\centering
\caption{Average Inference Time per Sample}
\label{tab:efficiency}
\small
\begin{tabular}{l|c|c}
\toprule
Stage & Time (s) & Percentage \\
\midrule
Stage 1: Contextual Retrieval & 3.42 & 17.67\% \\
Stage 2: Meta-Planning & 2.41 & 12.46\% \\
Stage 3: Multi-Agent Reasoning & 11.35 & 58.65\% \\
Stage 4: Synthesis \& Judgment & 2.17 & 11.21\% \\
\midrule
\textbf{Total} & \textbf{19.35} & \textbf{100\%} \\
\bottomrule
\end{tabular}
\end{table}

Multi-agent reasoning dominates inference time at 58.67\%, but the meta-planner mitigates this by selecting minimal agent ensembles. Simple Irony plans activate only 1-3 agents compared to 4-5 for Expectation Violation plans, achieving 35\% reduction in agent invocations for appropriate cases. Although the framework involves multiple reasoning stages, the adoption of parallel execution substantially improves runtime efficiency: processing 320 samples completes within approximately 20 minutes.

\subsection{Interpretability Quality Evaluation}
\begin{figure*}[!t]
    \centering
    \includegraphics[width=0.49\textwidth]{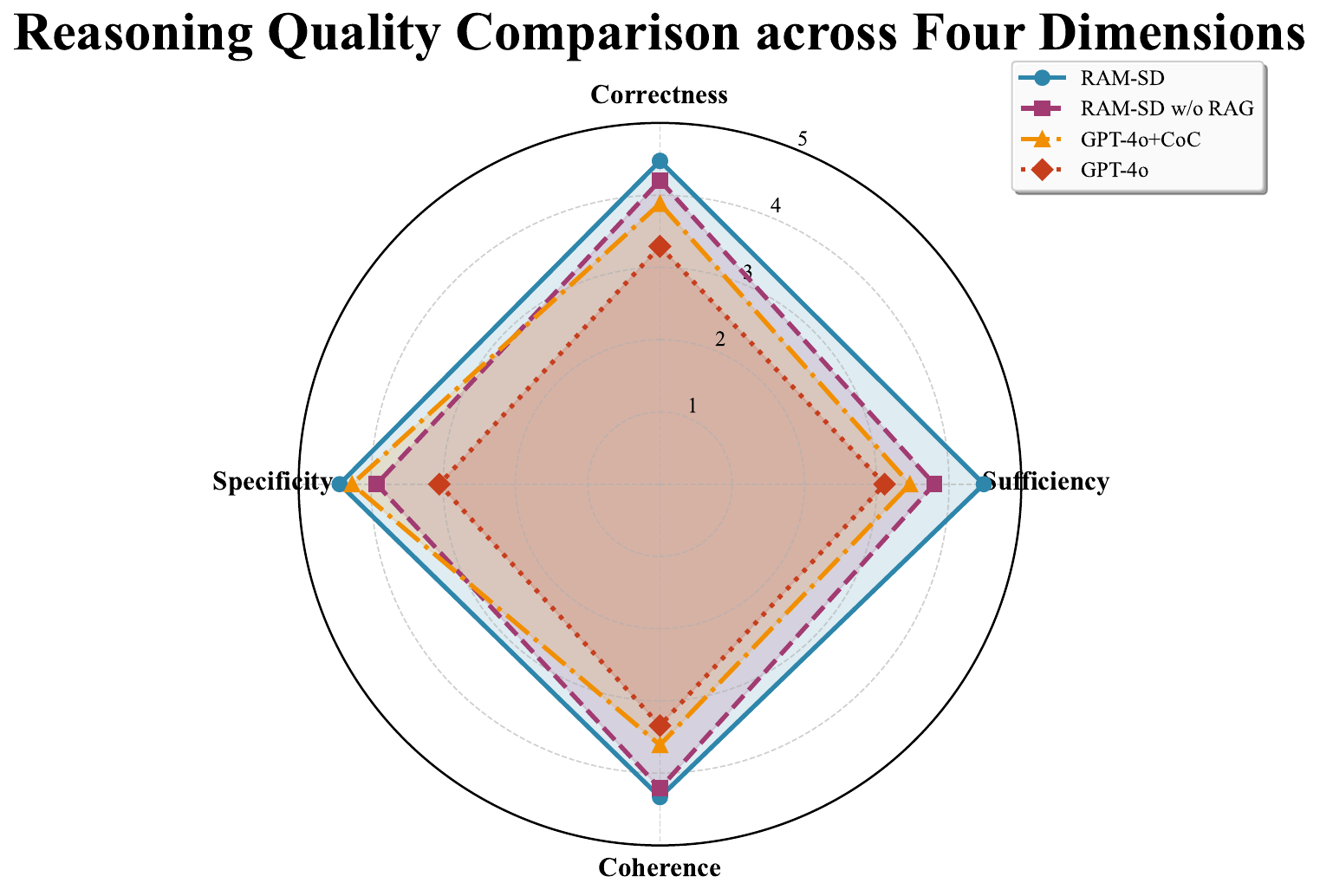}
    \hfill
    \includegraphics[width=0.50\textwidth]{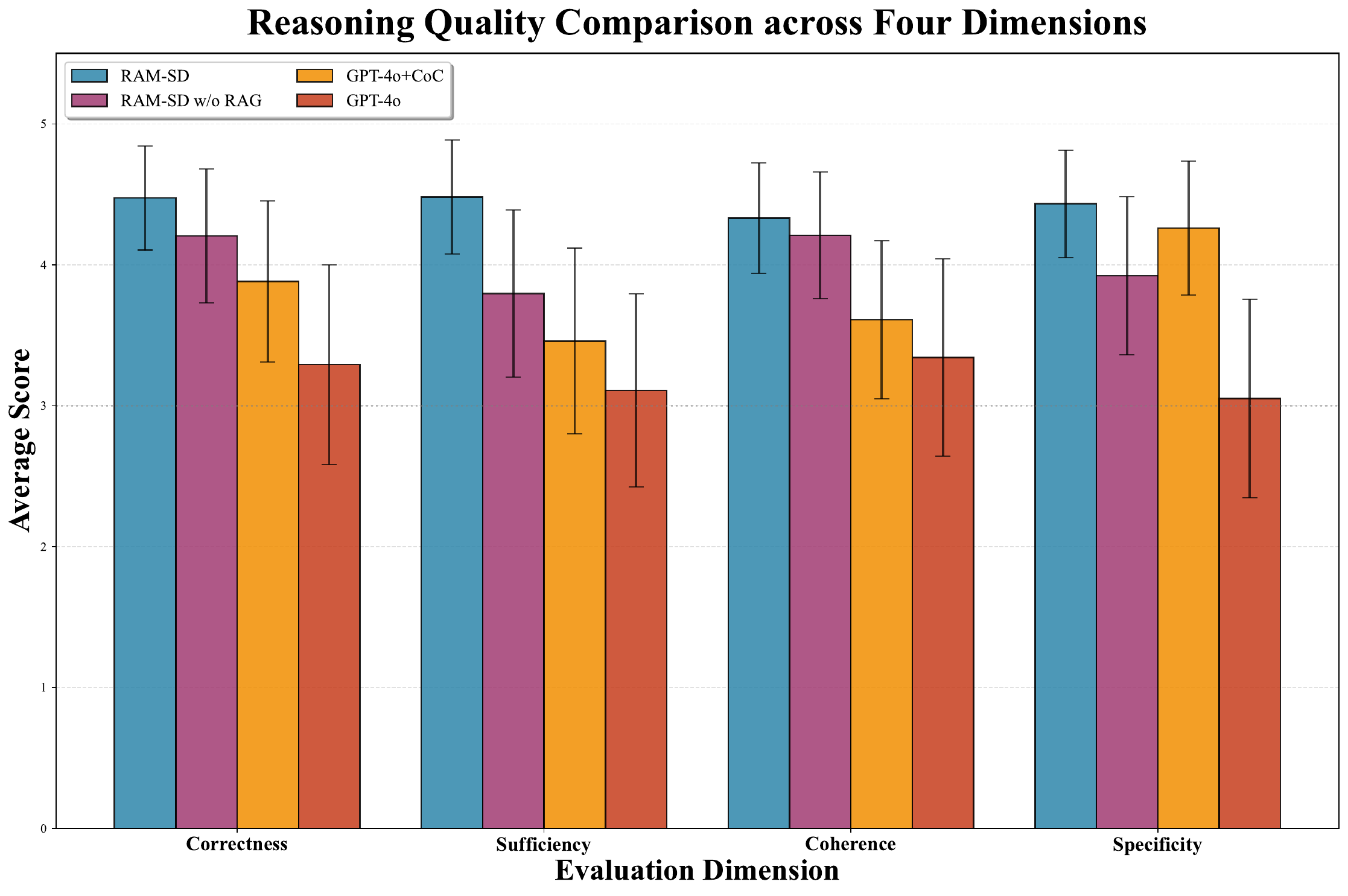}
    \caption{LLM-as-Judge evaluation results. \textbf{Left:} Radar chart across four dimensions, RAM-SD (blue) dominates, while GPT-4o+CoC (orange) achieves competitive Specificity, surpassing RAM-SD w/o RAG (purple). \textbf{Right:} Grouped bar chart showing mean scores with standard deviation bars for each dimension.}
    \label{fig:interpretability_comparison}
\end{figure*}

To validate that RAM-SD produces higher-quality explanations beyond accuracy gains, we employ \textbf{LLM-as-Judge} evaluation on 200 randomly sampled instances (50 per dataset, balanced) comparing four systems: \textbf{RAM-SD}, \textbf{RAM-SD w/o RAG}, \textbf{GPT-4o+CoC}, and \textbf{GPT-4o}. Using GPT-4o as judge, we evaluate reasoning traces along four key dimensions, \textbf{Correctness}, \textbf{Sufficiency}, \textbf{Coherence}, and \textbf{Specificity}, reflecting logical validity, completeness of evidence, internal consistency, and concreteness of expression.

\begin{table}[ht]
\centering
\caption{LLM-as-Judge Evaluation (n=200). Scores are reported on a 5-point Likert scale (5=best) for Correctness (Corr.), Sufficiency (Suff.), Coherence (Cohe.), and Specificity (Spec.). All improvements of RAM-SD over the baselines are statistically significant.}
\label{tab:interpretability}
\scriptsize
\begin{tabular*}{\columnwidth}{@{\extracolsep{\fill}}l|cccc|c}
\toprule
\textbf{System} & \textbf{Corr.} & \textbf{Suff.} & \textbf{Cohe.} & \textbf{Spec.} & \textbf{Overall} \\
\midrule
\textbf{RAM-SD} & \textbf{4.47} & \textbf{4.48} & \textbf{4.33} & \textbf{4.43} & \textbf{4.43} \\
RAM-SD w/o RAG & 4.20 & 3.80 & 4.21 & 3.92 & 4.03 \\
GPT-4o+CoC & 3.88 & 3.46 & 3.61 & 4.26 & 3.80 \\
GPT-4o & 3.29 & 3.11 & 3.34 & 3.05 & 3.20 \\
\bottomrule
\end{tabular*}
\end{table}

\textbf{Results.} As shown in Table~\ref{tab:interpretability}, RAM-SD achieves an overall score of 4.43, representing a significant improvement of 16.5\% over GPT-4o+CoC (score of 3.80) and 38.5\% over the standard GPT-4o (score of 3.20). Notably, while GPT-4o+CoC demonstrates highly competitive performance in \textbf{Specificity} with a score of 4.26 that approaches RAM-SD's 4.43, the advantages of our framework are most pronounced in \textbf{Sufficiency} and \textbf{Coherence}. In these dimensions, RAM-SD outperforms GPT-4o+CoC by 29.6\% and 20.0\% respectively, which we attribute to the richer evidence provided by the multi-agent architecture and the consistency enforced by the integrator. The ablation result for RAM-SD w/o RAG, which scored 4.03, confirms that the retrieval mechanism contributes a substantial 0.40 points (a 9.0\% relative contribution) to the final performance.

Figure~\ref{fig:interpretability_comparison} visualizes these results. The radar chart reveals an interesting pattern: while RAM-SD dominates most dimensions, GPT-4o+CoC achieves competitive Specificity scores, indicating calibration prompting's strength in producing concrete reasoning. The grouped bar chart with error bars clearly shows dimension-wise performance differences and variability across systems.

\textbf{Implications.} RAM-SD's structured reasoning enables transparent error diagnosis: users can inspect (1) retrieved exemplar relevance, (2) meta-planner strategy selection, (3) individual agent contributions, and (4) integrator synthesis. This interpretability is crucial for high-stakes deployments requiring human oversight.

\subsection{Error Analysis and Failure Patterns}
\subsubsection{Error Analysis}
\begin{figure}[t]
    \centering
    \includegraphics[width=\columnwidth]{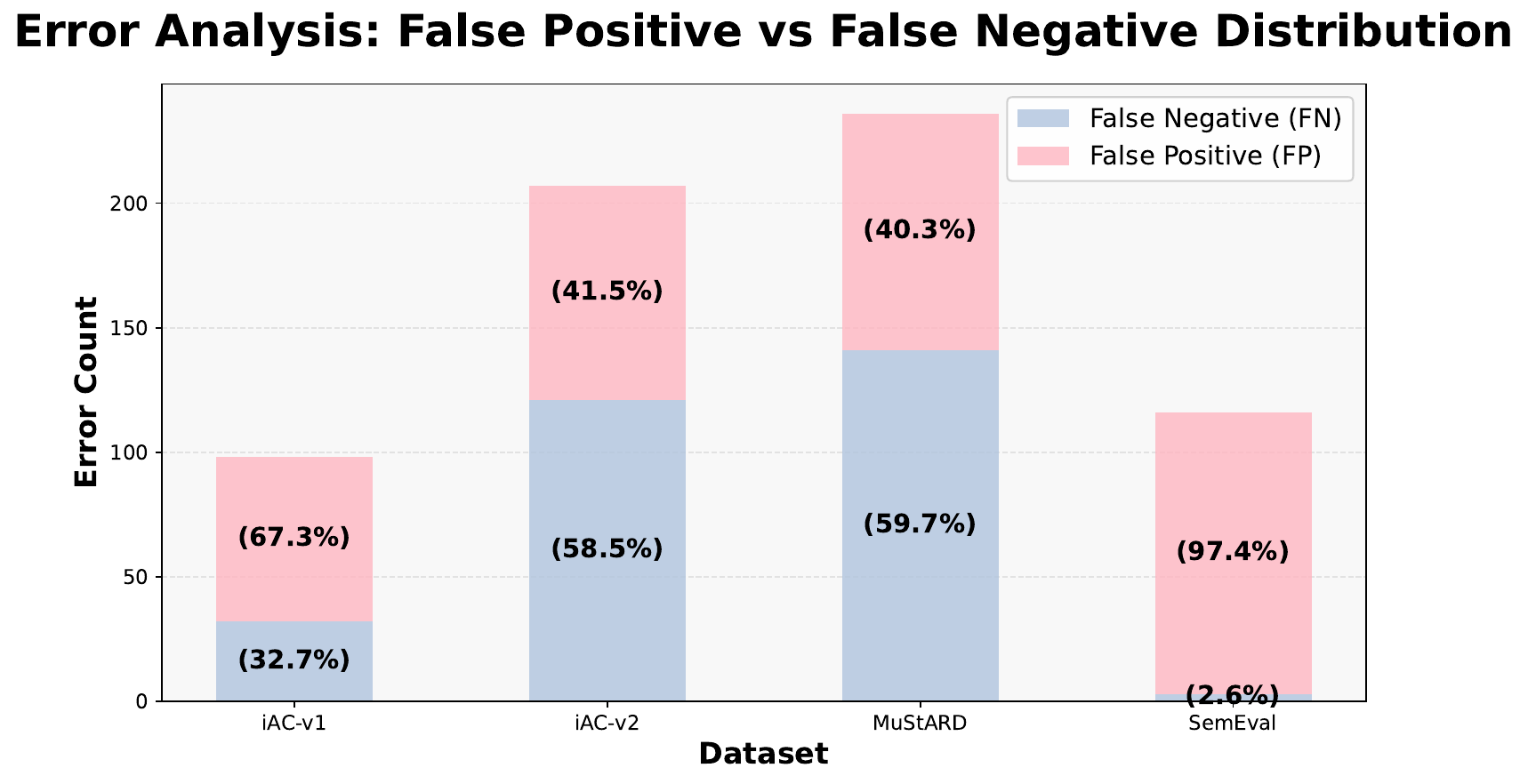}
    \caption{Error distribution showing distinct FP/FN patterns across datasets.}
    \label{fig:error_analysis}
\end{figure}

Figure~\ref{fig:error_analysis} shows error distributions across datasets, revealing retrieval-induced failure modes.

\textbf{Corpus Size Effects:} Our error analysis reveals a strong correlation between corpus characteristics and predominant error patterns, creating a dichotomy between datasets prone to over-prediction and those susceptible to under-detection. The IAC-v1 and SemEval datasets exhibit high false positive (FP) rates of 67.3\% and 97.4\%, respectively. This over-prediction in IAC-v1 is attributed to its small retrieval pool (1,595 samples), which forces a reliance on surface-level rhetorical markers that can misclassify figurative language. SemEval's FP bias stems from misleading structural cues, such as hashtags and @mentions, that superficially mimic sarcasm without the same pragmatic intent.

Conversely, the IAC-v2 and MUSTARD datasets show significant under-detection, with false negative (FN) rates of 58.5\% and 59.7\%. The larger corpus of IAC-v2 appears to induce a more conservative model behavior, as the abundance of diverse retrieved contexts dilutes confidence in more nuanced cases. Finally, MUSTARD's FN dominance highlights the inherent limitation of a text-only analysis on multimodal data, where essential audio-visual cues like vocal tone and facial expressions are fundamentally inaccessible.

\subsubsection{Case Study: Cascading Failure from Retrieval Bias}

We examine a representative SemEval FP error: \textit{``when you refer to yourself in the plural, you'll get that @RBRNetwork1...''} (Label: 0, Predicted: 1, Conf.: 0.85). Table~\ref{tab:case_study} traces the cascading failure.

\begin{table}[ht]
\centering
\caption{Stage-wise error propagation analysis.}
\label{tab:case_study}
\scriptsize
\begin{tabular}{p{1.8cm}p{5.7cm}}
\toprule
\textbf{Stage} & \textbf{Output \& Failure Mechanism} \\
\midrule
\textbf{Retrieval} & Retrieved: \textit{``LOOL from guy with multiple handles''}, \textit{``9 Followers, now relevant?''}
\newline \textbf{Issue:} Matched Twitter structure (@mentions), not pragmatic intent (mockery vs. meta-commentary). \\
\midrule
\textbf{Meta-Plan} & Plan: $\mathcal{P}_{\text{EV}}$; $O_{\text{plan}}$: ``indirect mockery''
\newline \textbf{Issue:} Biased priming from retrieval. \\
\midrule
\textbf{Agents} & $A_{\text{rhet}}$: ``mild irony''; $A_{\text{incon}}$: 6/10; $A_{\text{exp}}$: ``mocks plural self-ref''
\newline \textbf{Issue:} Weak signals amplified by biased $O_{\text{plan}}$. \\
\midrule
\textbf{Synthesis} & Judger: $p=0.85$; ``strong sarcastic alignment''
\newline \textbf{Issue:} No mechanism to challenge consensus. \\
\bottomrule
\end{tabular}
\end{table}

This exemplifies the \textit{echo chamber effect}: biased retrieval constrains meta-planning, which primes agents toward confirmation, yielding high-confidence errors. Our symbolic pipeline lacks gradient-based error attribution, making such cascades hard to detect. Mitigation requires \textit{contrastive retrieval} and \textit{agent-level calibration} to challenge weak unanimous signals.

\section{Conclusion}
We introduced RAM-SD, a novel four-stage multi-agent framework for sarcasm detection. Its core meta-planner dynamically orchestrates specialized agents based on retrieval-augmented analysis of the input text. The framework achieves a state-of-the-art 77.74\% average Macro-F1, improving upon the GPT-4o+CoC baseline by 7.01 points. Ablation studies validate this hierarchical design, confirming the meta-planner and retrieval module are critical components that contribute 3.27 and 2.69 percent to the Macro-F1 score respectively. Our results demonstrate the efficacy of structured multi-agent systems for nuanced language understanding. We position RAM-SD as an extensible, backward-compatible paradigm: practitioners can plug in new plans/agents as domains evolve. We expect this line of work to guide a shift from opaque single-pass classifiers to context-grounded, plan-conditioned reasoning frameworks.
\newpage

\section*{Limitations}

  \textbf{Fundamental Limitations of Retrieval-Based Reasoning.}
  Our framework inherits the cold-start problem of case-based systems:
  performance degrades when test instances lack precedent in the retrieval
  corpus. MuStARD's FN errors exemplify this—text retrieval cannot ground
  multimodal sarcasm requiring audio-visual cues. More critically, the
  system struggles with emergent or novel sarcasm forms (e.g., new memes,
  evolving cultural references) that fall outside observed patterns. This
  suggests retrieval alone is insufficient; compositional reasoning or
  generative world models are needed for genuine generalization.

  \textbf{Agent Coordination and Conflict Resolution.}
  While the Integrator synthesizes agent outputs, our framework lacks
  principled mechanisms for contradictory evidence (e.g., when Rhetoric
  Agent detects hyperbole but Knowledge Agent finds factual consistency).
  Current reliance on implicit weighting by the Judger is brittle. Future
  work should incorporate confidence calibration, cross-agent consistency
  checks, or argumentation-theoretic arbitration protocols.





\bibliography{acl_references}

\appendix


\section{Appendix}
\label{sec:prompts}

This appendix provides the complete prompt templates used in each stage of the RAM-SD framework. All prompts are executed using GPT-4o with temperature=0.1 for consistency.

\definecolor{PromptShell}{RGB}{248,249,250}
\definecolor{PromptFrame}{RGB}{108,117,125}
\definecolor{ModifiedPrompt}{RGB}{255,243,243}
\definecolor{NewPrompt}{RGB}{243,250,255}

\tcbset{
  promptbox/.style={
    breakable,
    width=\linewidth,
    boxrule=0.6pt,
    colframe=PromptFrame,
    colback=PromptShell,
    coltitle=white,
    colbacktitle=PromptFrame,
    left=3pt,right=3pt,top=3pt,bottom=3pt,
    fonttitle=\small\bfseries,
    fontupper=\ttfamily\scriptsize,
    before skip=8pt,
    after skip=8pt,
  },
  modifiedprompt/.style={
    promptbox,
    colback=ModifiedPrompt,
  },
  newprompt/.style={
    promptbox,
    colback=NewPrompt,
  }
}

\subsection{Stage 1: Contextual Retrieval}

\subsubsection{Step 1.1: Rationale Generation Prompt}
\label{prompt:rationale}

This prompt generates explanatory rationales for each retrieved exemplar to create rationale-augmented context $\mathcal{C}_{\text{aug}}$.

\begin{tcolorbox}[promptbox, title=Rationale Generator]
Analyze why the following text is [sarcastic/non-sarcastic]:

Text: "\{text\}"
Label: \{label\}

Provide a concise rationale (2-3 sentences) explaining:
1. Key linguistic features (tone, style, rhetorical devices)
2. Contextual cues or knowledge dependencies
3. Why this is clearly [sarcastic/non-sarcastic]

Focus on concrete evidence rather than abstract concepts.
\end{tcolorbox}

\textbf{Note:} This prompt is applied to all $2k$ retrieved exemplars to form the rationale-augmented context set $\mathcal{C}_{\text{aug}} = \{(T_i, y_i, r_i)\}$.

\subsubsection{Step 1.2: Similarity Analysis Prompt}
\label{prompt:similarity_analysis}

 After generating rationales, this prompt analyzes the query's similarity patterns with both sarcastic and non-sarcastic exemplars. The analysis output informs the Meta-Planner in Stage 2.

\begin{tcolorbox}[newprompt, title=Similarity Analyzer, fontupper=\ttfamily\scriptsize]
You are a similarity analysis expert. Compare the query text with retrieved
exemplars to identify linguistic patterns and stylistic similarities.

**Query Text:** "\{text\}"
\{context\_if\_available\}

**Retrieved NON-SARCASTIC Examples (Label 0) with Rationales:**
1. Text: "\{example\_1\}" | Rationale: \{rationale\_1\}
2. Text: "\{example\_2\}" | Rationale: \{rationale\_2\}
3. Text: "\{example\_3\}" | Rationale: \{rationale\_3\}

**Retrieved SARCASTIC Examples (Label 1) with Rationales:**
1. Text: "\{example\_1\}" | Rationale: \{rationale\_1\}
2. Text: "\{example\_2\}" | Rationale: \{rationale\_2\}
3. Text: "\{example\_3\}" | Rationale: \{rationale\_3\}

Analyze the query's similarity patterns:

**Similarity to NON-SARCASTIC Examples:**
- Shared features: [tone, directness, literal expression, etc.]
- Pattern alignment: [similar linguistic structures or rhetorical styles]
- Strength of similarity: [strong/moderate/weak]

**Similarity to SARCASTIC Examples:**
- Shared features: [irony markers, contradiction patterns, mocking tone, etc.]
- Pattern alignment: [similar ironic devices or expectation violations]
- Strength of similarity: [strong/moderate/weak]

**Comparative Assessment:**
- Primary similarity direction: [more similar to sarcastic/non-sarcastic/mixed]
- Key discriminative features: [2-3 critical differences from one category]
- Confidence level: [high/medium/low]

**Contextual Inference:**
Based on retrieved examples and their rationales, infer:
- Likely situational context of the query
- Potential background knowledge required for interpretation
- Candidate sarcasm type if sarcastic: [expectation\_violation/knowledge\_dependent/
  simple\_irony]

Return structured analysis focusing on concrete pattern matching with exemplars.
\end{tcolorbox}

\textbf{Output:} This produces similarity analysis $S_{\text{analysis}}$ that captures pattern-matching insights between query and exemplars, which is passed to Meta-Planner along with $\mathcal{C}_{\text{aug}}$.

\subsection{Stage 2: Retrieval-Augmented Meta-Planning}

\subsubsection{Meta-Planner Prompt}
\label{prompt:metaplanner}

 This prompt integrates similarity analysis from Stage 1 with feature analysis to select optimal reasoning plan. Produces both plan selection ($P^{*}$) and contextual analysis ($O_{\text{plan}}$).

\begin{tcolorbox}[modifiedprompt, title=Meta-Planner (Retrieval-Augmented), fontupper=\ttfamily\tiny]
As meta-planner, analyze query and select optimal reasoning strategy.

**Query:** "\{text\}"

**Similarity Analysis from Stage 1:**
\{similarity\_analysis\_output\}

**Retrieved Examples Summary:**
- Non-sarcastic: \{brief\_summary\_of\_non\_sarc\_examples\}
- Sarcastic: \{brief\_summary\_of\_sarc\_examples\}

Based on similarity patterns and query features, analyze:

**Feature Analysis:**
1. Contradiction level: [none/low/medium/high] - between literal/intended meaning
2. Exaggeration/irony present: [yes/no]
3. Emotional conflict: [yes/no] - positive words with negative intent
4. Context dependency: [low/medium/high]
5. Rhetorical devices: [list key devices if present]
6. Knowledge entities: [list SPECIFIC entities: named persons, events, orgs]
   Exclude: generic concepts, common terms, abstract ideas

**Plan Selection (choose ONE):**
- expectation\_violation: For contradiction/irony, pragmatic violations
- knowledge\_dependent: For SPECIFIC entities requiring background knowledge
- simple\_irony: For short (fewer than 15 words), overt ironic expressions

**Contextual Analysis (O\_plan):**
Synthesize inferred context from similarity analysis and examples:
- Likely situational context: [inferred scenario/setting]
- Relevant background knowledge: [key info for understanding]
- Expectation baseline: [what would be normal expression in this context]
- Pragmatic interpretation hints: [guidance for downstream agents]

**Output:**
\{"selected\_plan": "expectation\_violation/knowledge\_dependent/simple\_irony",
 "confidence": 0-1,
 "contextual\_analysis": "synthesized context and interpretation guidance",
 "reasoning": "why this plan based on similarity patterns and features"\}

Priority: Weight similarity patterns heavily - if query strongly matches sarcastic/
non-sarcastic exemplar patterns, factor this into plan selection.
\end{tcolorbox}

\textbf{Output:} Produces $(P^{*}, O_{\text{plan}})$ where $P^{*}$ determines agent ensemble for Stage 3, and $O_{\text{plan}}$ provides contextual grounding for all agents.

\textbf{Implementation Note:} LLM selection validated by rules (text length greater than 50 words overrides \texttt{simple\_irony}; 3+ specific entities triggers \texttt{knowledge\_dependent}).

\subsection{Stage 3: Agent-Based Multi-View Reasoning}

All specialized agents receive the query text $T_q$, contextual analysis $O_{\text{plan}}$ from Stage~2, and can reference the similarity analysis $S_{\text{analysis}}$ and rationale-augmented exemplars $\mathcal{C}_{\text{aug}}$ when needed.







\subsubsection{Semantic Agent Prompt}
\label{prompt:semantic}

This agent ($A_{\text{sem}}$) performs semantic analysis.

\begin{tcolorbox}[promptbox, title=Semantic Agent]
As a semantic analysis expert, deeply analyze the semantics of the text:

Text: "\{text\}"
Context: "\{context\}" (if available)

Analysis:
1. Literal meaning and deep meaning
2. Emotional tendency and intensity (positive/negative/neutral, intensity 1-10)
3. Emotional coloring of key vocabulary
4. Overall tone (formal/informal/teasing/serious, etc.)

Return precise, concise structured analysis in several sentences.
\end{tcolorbox}

\subsubsection{Expectation Agent Prompt}
\label{prompt:expectation}

This agent ($A_{\text{exp}}$) builds expectation models using contextual grounding from Meta-Planner.

\begin{tcolorbox}[promptbox, title=Expectation Agent]
Build expectation model using prior analysis:

Text: "\{text\}"

Analysis from prior agents:
- Context analysis (from Context Agent): \{context\_agent\_output\}
- Semantic analysis (from Semantic Agent): \{semantic\_agent\_output\}
- Contextual grounding (from Meta-Planner): \{meta\_planner\_context\}

Build expectation model:
1. Normal expectation: What would be typical expression in this context
2. Actual vs. expected: Compare query expression with baseline expectation
3. Deviation analysis: Degree (1-10) and type (semantic/tonal/pragmatic)
4. Intentionality: Whether deviation appears deliberate for rhetorical effect

Reference Meta-Planner's expectation baseline for contextual grounding.
Return structured analysis (3-4 sentences).
\end{tcolorbox}

\subsubsection{Knowledge Agent Prompt}
\label{prompt:knowledge}

This agent ($A_{\text{know}}$) retrieves background knowledge.

\begin{tcolorbox}[promptbox, title=Knowledge Agent]
As a knowledge retrieval expert, analyze entities and concepts in the text:

Text: "\{text\}"
Context: \{context\_agent\_output\}

Please retrieve and analyze:
1. Key entities, people, events mentioned in the text
2. Common evaluations, stereotypes, public perceptions of these entities
3. Related background knowledge and common sense
4. Importance of this knowledge for understanding text sarcasm

Return precise, concise structured analysis in several sentences.
\end{tcolorbox}

\subsubsection{Alignment Agent Prompt}
\label{prompt:alignment}

This agent ($A_{\text{align}}$) checks alignment between text and knowledge, referencing similarity patterns from Stage 1.

\begin{tcolorbox}[promptbox, title=Alignment Agent]
Check alignment between text and background knowledge:

Text: "\{text\}"
Semantic analysis: \{semantic\_agent\_output\}
Background knowledge: \{knowledge\_agent\_output\}
Similarity patterns (from Stage 1): \{similarity\_analysis\_summary\}

Analyze:
1. Consistency: Does literal meaning align with background knowledge/common sense
2. Distortion detection: Deliberate misrepresentation or counter-factual claims
3. Inconsistency assessment: Degree (1-10) and nature (factual/evaluative/tonal)
4. Sarcastic intent: Whether inconsistency serves ironic/mocking purpose

Consider similarity patterns - if query matches known sarcastic patterns with
knowledge contradictions, weight this evidence. Return structured analysis
(3-4 sentences).
\end{tcolorbox}

\subsubsection{Incongruity Agent Prompt}
\label{prompt:incongruity}

This agent ($A_{\text{incon}}$) detects inconsistencies.

\begin{tcolorbox}[promptbox, title=Incongruity Agent]
Specifically detect and quantify inconsistencies:

Text: "\{text\}"
Expectation analysis: \{expectation\_agent\_output\}

Please detect:
1. Specific types of inconsistency (semantic, emotional, logical, common sense, etc.)
2. Quantification of inconsistency degree (1-10)
3. Whether inconsistency has sarcastic effect
4. Key inconsistency points location and manifestation

Return precise, concise structured analysis in several sentences.
\end{tcolorbox}

\subsubsection{Rhetoric Agent Prompt}
\label{prompt:rhetoric}

This agent ($A_{\text{rhet}}$) identifies rhetorical devices. Added constraints to reduce false positives from over-interpretation.

\begin{tcolorbox}[modifiedprompt, title=Rhetoric Agent]
Identify if there is any rhetorical devices in the text:

Text: "\{text\}"
Context: "\{context\}" (if available)

Focus on identifying:
1. Irony, exaggeration, understatement
2. Puns, rhetorical questions, interrogative sentences
3. Contrast, analogy
4. Contribution of these rhetorical devices to the expression
5. Do not think too much, these may be replies to other people on social media
6. Judge it as you are a normal linguist, not a sarcasm detector

Return precise, concise structured analysis in several sentences.
\end{tcolorbox}

\textbf{Modification Rationale:} Points 5-6 instruct the agent to avoid over-analyzing casual social media language, reducing false positive rate (see Section 4.7.1 error analysis).

\subsection{Stage 4: Synthesis and Final Judgment}

\subsubsection{Integrator  Prompt}
\label{prompt:integrator}

Enhanced with explicit decision logic, strict standards, and integration of Stage 1 similarity analysis to address over-prediction bias.

\begin{tcolorbox}[modifiedprompt, title=Integrator, fontupper=\ttfamily\tiny]
As sarcasm expert, synthesize all evidence and make final judgment:

**Query:** "\{text\}"

**Similarity Analysis (Stage 1):**
\{similarity\_analysis\_from\_stage1\}

**Meta-Planner Output (Stage 2):**
Selected plan: \{plan\_type\}, Contextual analysis: \{O\_plan\_summary\}

**Agent Outputs (Stage 3):**
\{all\_agent\_outputs\_json\}

**Decision Framework:**

*Positive Indicators:* Contradiction literal/intended; rhetorical questions with
mocking; exaggeration/understatement; emotional incongruity; ironic praise;
expectation violations; strong similarity to sarcastic exemplars

*Negative Indicators:* Direct criticism without irony; genuine questions;
consistent tone; literal matches intent; strong similarity to non-sarcastic
exemplars

*Synthesis Requirements:*
1. Weight similarity analysis heavily - if Stage 1 shows strong pattern match to
   sarcastic/non-sarcastic, prioritize this evidence
2. Evaluate coherence: Do multiple agents agree on sarcastic interpretation?
3. Assess intentionality: Are contradictions/devices deliberate rhetorical strategy?
4. Context alignment: Does judgment match Meta-Planner's contextual inference?

**STRICT Decision Logic:**
- Multiple clear indicators (irony+contradiction+mocking) + similarity to sarcastic
  → sarcastic
- Strong similarity to sarcastic patterns + agent consensus on irony → sarcastic
- Obvious contradiction literal/intended with intentionality → sarcastic
- Direct criticism/opinions without ironic devices → non-sarcastic
- Genuine inquiry (not mockery) + similarity to non-sarcastic → non-sarcastic
- Emotional expression without contradiction + non-sarcastic patterns → non-sarcastic

**Domain Context:** Forum posts/comments/replies. Apply STRICT standards - only
classify sarcastic with CLEAR evidence (intentional irony/mockery/contradiction).
Casual criticism without ironic intent → non-sarcastic.

**Output Format:**
Line 1: <<LABEL>> 1 (sarcastic) or 0 (non-sarcastic)
Line 2: JSON \{"label": 1/0, "conf": 0-1, "reasoning": "synthesis referencing
similarity analysis, agent findings, and decision rationale"\}
\end{tcolorbox}

\end{document}